\documentclass{article}

\usepackage[nonatbib,preprint]{neurips_2025}

\usepackage[utf8]{inputenc} 
\usepackage[T1]{fontenc}    
\usepackage{hyperref}       
\usepackage{url}            
\usepackage{booktabs}       
\usepackage{amsfonts}       
\usepackage{nicefrac}       
\usepackage{microtype}      
\usepackage{xcolor}         
\usepackage{graphicx}

\title{SHARDeg: A Benchmark for Skeletal Human Action Recognition in Degraded Scenarios}

\author{Simon Malzard\\ 
The Alan Turing Institute\\
The Alan Turing Institute, London NW1 2DB, UK.\\
\and
Nitish Mital\\ 
The Alan Turing Institute\\
The Alan Turing Institute, London NW1 2DB, UK.\\
\and
Richard Walters\\ 
The Alan Turing Institute\\
The Alan Turing Institute, London NW1 2DB, UK.\\
\and 
Victoria Nockles\\ 
The Alan Turing Institute\\
The Alan Turing Institute, London NW1 2DB, UK.\\
\and
Raghuveer Rao\\
DEVCOM Army Research Laboratory, USA\\
\and
Celso M. De Melo\\
DEVCOM Army Research Laboratory, USA\\
}

\begin{document}

\maketitle
\begin{abstract}

Computer vision models for detection, prediction or classification tasks operate on video data-streams that are often degraded in the real world, due to deployment in real-time or on resource-constrained hardware. It is therefore critical that these models are robust to degraded data, but state of the art (SoTA) models are often insufficiently assessed with these real-world constraints in mind. This issue is exemplified by Skeletal Human Action Recognition (SHAR), which is a critical element in many computer vision pipelines operating in real-time and at the edge, but where robustness to degraded data has previously only been shallowly and inconsistently assessed. Here we address this issue for SHAR by providing an important first data degradation benchmark on the most detailed and largest 3D open dataset, NTU-RGB+D-120, and assess the robustness of five leading SHAR models to three forms of degradation that are representative of real-world issues. We demonstrate the need for this benchmark by showing that the form of degradation, which has not previously been considered, has a large impact on model accuracy; at the same effective frame rate, model accuracy can vary by $>40\%$ depending on degradation type. We also identify that temporal regularity of frames in degraded SHAR data is likely a major driver of differences in model performance, and harness this information to improve performance of existing models by up to $>40\%$, through employing a simple mitigation approach based on interpolation. Finally, we highlight how our benchmark has helped identify an important degradation-resistant SHAR model based in Rough Path Theory; the LogSigRNN SHAR model outperforms the SoTA DeGCN model in five out of six cases at low frame rates (3 FPS, common for real-world deployments), by an average accuracy of $6\%$, despite trailing the SoTA model by $11-12\%$ on un-degraded data at high frame rates (30 FPS). Our new benchmark will support the community’s ability to assess, improve and develop new SHAR models and ensure their suitability for practical real-world deployment. Furthermore, broader adoption of our systematic approach to assessing model performance on purposely-degraded video datasets will foster the development of a new generation of degradation-resilient models across many computer vision domains.

\end{abstract}    
\section{Introduction}

In the real-world, computer vision models for detection, prediction or classification tasks commonly operate on video data-streams that have degraded sampling-rates relative to idealised training datasets \cite{Liu2024}. Constraints or limitations associated with real-world deployment are a major source of this degradation; real-time operation or resource-constrained hardware commonly lead to loss of frames, either by design (e.g. video compression, dynamic changes in capture rates \cite{Ma2020}) or unintentionally (e.g. bandwidth/latency problems \cite{Goel2020}, hardware failures \cite{Qutub2022} or differences between network or recording equipment).

It is therefore critical that computer vision models are robust to degraded video-streams, especially those that are intended for deployment in edge, real-time or other challenging conditions. However, such robustness requires careful and systematic evaluation of models under the conditions in which they will be typically deployed. But state-of-the-art (SoTA) models are primarily benchmarked against well-known high-quality idealised datasets, and performance on degraded video is often represented only through insufficient or non-standardised ablation studies (e.g. \cite{Michaelis2019}), or else neglected completely (e.g. \cite{Hendrycks2019}). This can therefore mean that published model rankings may be misleading for practical applications; for example, models that are considered SoTA may actually have worse performance than competing models under common real-world conditions.

This issue is exemplified by Skeletal Human Action Recognition (SHAR), an important computer vision subfield that involves classification of human actions from video-derived skeletal time-series. Classification of human actions is a critical element of many practical computer vision pipelines that involve interacting with humans or identifying human intentions, including human-robot teaming \cite{Mehak2024}, violence detection \cite{Nardelli2024}, video surveillance \cite{Kim2024}, assisted living and smart home systems \cite{Bibbo2023}, motion rehabilitation and sports monitoring \cite{Zhang2024}. The majority of these applications are intended for real-time or edge use, and SHAR is designed to support these deployment considerations. Skeletal segmentation of video data reduces dimensionality and computational complexity, allowing for more robust and compact representation of human motion, and use of lighter-weight, graph-based models for classifying actions. Skeletal segmentation involves obtaining key skeletal markers associated with individual joints or bones. This is commonly achieved through hand-labelling video data or via wearable motion capture sensors or infrared depth cameras, and increasingly through use of pose-estimation tools such as OpenPose \cite{Cao2019}, which extract skeletal time series directly from RGB video data.
But whilst the applications of SHAR and the extraction of skeletal time-series reflect the intended use-case at the edge, SHAR research has largely focused on data from well-controlled laboratory environments, assuming high frame-rate and reliable delivery of frames. Ablation studies that assess the relative impact of deviations from these idealised conditions have been under explored, and with little standardisation. Most action classification models are trained on data with frame-rates of 30 frames per second (FPS), but real-world cameras and CCTV units in ideal conditions have capture rates which typically range from 10-30 and 5-15 FPS \cite{Arn2016}, respectively. These frame rates can be even lower due to compression, e.g. to reduce storage and processing costs \cite{Gowda2021}. Indeed, it has been shown that real-world pipelines using SoTA SHAR models are only capable of processing at 3 FPS due to excessive and fluctuating latency \cite{Wang2024}. And it has been shown that human action recognition models in general have reduced performance when inference is run on data with lower frame rates than training examples \cite{Anil2024}. 

Therefore, to facilitate the use of light-weight SHAR models in real-world edge environments, we need benchmarks for assessing the performance of models under different operating conditions. This will enable the community both to understand the shortcomings of existing models and to encourage the building of a new generation of more robust models. In this study we propose to address this problem through careful and systematic degrading of data, which enables us to measure model performance across a breadth of real-world scenarios where train-time setups do not match test-time setups. Before laying out our aims in detail, we first justify our specific selections of dataset, degradation methods, and models to test, with respect to previous related research.

\subsection{Related Work}

In this section we first discuss SHAR datasets and previous research focused on degraded SHAR data, and then outline the most important spatial-temporal learning architectures for SHAR to explain our selection of five models for our benchmark.  

\noindent{\bf{Datasets:}} Skeletal HAR datasets in the open domain vary widely in terms of the source of the recordings, settings, volume of classes, participants, view-points, video clips per class, diversity of classes and length of video sequences. Datasets also contain different skeletal segmentation sources such as hand annotations, pose estimations and depth camera measurements. Segmentations vary in the number of joints and either contain 2D or 3D coordinates, where more coordinates and more joints help to further discriminate between different action classes. For example, the HMDB51 \cite{HMDB51} dataset contains $7000$ videos with $51$ action classes taken from various online sources including movies and web videos, containing 2D segmentations with $14$ joints from 2D pose estimation. In comparison the Kinetics \cite{Kinetics} dataset contains substantially more videos and classes, with $300,000$ videos sourced from YouTube clips of roughly 10s in length, covering 400 action classes. This dataset also uses 2D segmentations with $18$ joints derived from 2D pose estimation. However, these datasets are sourced from the Web and contain inbuilt biases and confounding factors derived from complex backgrounds and lower segmentation qualities. Furthermore, with the advancement in SoTA pose estimation technologies such as OpenPose to generate 3D segmentations, these datasets do not represent the SoTA for SHAR. In contrast, NTU-RGB+D-120 \cite{Liu2019} is a comprehensive, high quality dataset, captured in controlled settings containing $114,480$ video clips evenly spread across $120$ classes, involving single single person, person and object and person-to-person interactions. Skeletal segmentations are derived from Microsoft Kinect V2 sensor infrared depth cameras which provide 3D segmentations with 25 joints. The larger number of joints, quality of the segmentations and less inbuilt biases and confounding factors makes NTU-RGB+D-120 an ideal candidate for a first benchmark for assessing the impact of temporal data degradation on SHAR; the dataset represents a carefully controlled and high quality baseline against which we can accurately measure the impact of degradation. 



\noindent{\bf{The impact of degraded data-streams on Human Action Recognition:}} The first studies that assessed the impact of video quality and frame rates on action recognition focused on video rather than Skeletal HAR, and identified that SoTA models were not robust to variations in frame rates \cite{Harjanto2016,See2015}. These studies used this as motivation to design key-frame selection methods \cite{Harjanto2016} and Spatio-Temporal Interest Points \cite{See2015}. Extraction of key frames has remained the main focus for investigations into frame rate changes and sampling for video HAR (e.g. \cite{Korbar2019}, \cite{Wu2019}, \cite{zhu2020}), and most studies consider either uniform subsampling where frames are selected at regular intervals, or dense subsampling where frames are selected either uniformly or randomly at much higher frequencies. 

However, for SHAR the impact of frame rate has not been studied systematically, and comparisons between models are rare. Random subsampling is the most common type of subsampling that is investigated, predominantly as part of ablation studies to check the impact of frame rate on performance (e.g. \cite{Yang2022} \cite{Liao2021}). But random subsampling is not representative of the range of realistic subsampling modalities. More careful exploration of this range would give a more accurate reflection of model robustness (and variability in robustness) in real world scenarios. 

To first order, variation in time-series degradation can be considered as a spectrum of increasing regularity of frame dropout. Benchmarks containing spatial-temporal corruptions in CNNs and Transformers in the CV domain \cite{Yi2021} have explored this spectrum which ranges from total occlusion from fog to shot noise and packet loss (similar to random subsampling) all the way to compression and frame rate conversions (e.g. uniform sampling). For our SHAR benchmark we use the end members of this regularity spectrum in addition to the commonly-used random subsampling as our three degradation types. The two end members represent the effects of whole-skeleton occlusion (e.g. from fog) and variations in frame rates caused by uniform degradation between training and test time on SHAR.  These common forms of degradation are the result of camera effects and malfunctions upstream of the inference and skeletal segmentation processes. We consider pose-estimation effects to be second order to camera malfunctions \cite{Wang2021} and therefore beyond the scope of this work. 





\noindent{\bf{Skeletal time-series architectures:}} 
Early SHAR models used hand-crafted features to capture the dynamics of joint motion \cite{Ofli2014,Vemulapalli2014}, Recurrent Neural Networks (RNNs) for capturing long-term dynamics \cite{Lev2016,Zhang2017}, or Long-Short Term Memory (LSTM) networks to capture the kinetic dependency of correlations between adjacent joints \cite{Liu2016}. However, graph neural networks, and in particular the graph convolutional network (GCN) architecture, have emerged in recent years as the basis for most SoTA models in SHAR, due to their ability to capture longer distance spatial correlations between non-adjacent joints, and to provide a more natural representation of the spatial and temporal dynamics of skeletal motion \cite{Ren2024}.   

Most SHAR GCN models use a joint-based backbone, representing joints as graph vertices and temporal connections between body parts as graph edges (e.g. \cite{Yan2018STGCN}). But GCN models also commonly train on ensembles of different graph backbones such as joints, bones \cite{Shi2019}, joint-motion and bone-motion graphs \cite{Shi2020} or joint-bone fusion graphs \cite{Tu2022}. Bone graphs incorporate information such as bone length and orientation to reduce viewpoint sensitivity, whilst motion graphs are velocity aware. GCN-LSTM and GCN-Transformer hybrid architectures have also been proposed \cite{Do2024} with the aim of better learning the long-term spatial and temporal patterns of human motion. 

In this study we consider five different GCN-based SHAR models. MS-G3D Net \cite{Liu2020}, FR-HEAD \cite{Zhou2023} and DeGCN \cite{Myung2024} are models selected for having current (DeGCN) or previous (MS-G3D Net, FR-HEAD) SoTA performance on the SHAR task, and each represent a significant step in the recent evolution of GCNs for SHAR:
\begin{itemize}
    \item MS-G3DNet (MIT license) \cite{Liu2020} is a multi-scale 3D graph convolutional network which can disentangle multi-scale aggregation; this acts to avoid the dominance of joints from local body parts, which was an issue for earlier GCN SHAR models (e.g. \cite{Yan2018STGCN}).
    \item FR-HEAD (No license supplied with publication) \cite{Zhou2023} uses an auxiliary feature refinement head to obtain samples for difficult-to-distinguish actions (e.g. reading, writing and typing), and then uses contrastive learning to improve performance on these actions. This is in contrast to most previous SHAR ML models, which show lower performance on such actions. 
    \item DeGCN (MIT license) \cite{Myung2024} is the current SoTA SHAR model and uses an alternative method for improving performance on difficult-to-distinguish actions. This involves refining the graph representation for each action by adaptively selecting the most informative joints that can identify subtle differences between similar actions.
\end{itemize}

The last two models we consider are LogSigRNN \cite{Liao2021} and GCN-Dev-LSTM \cite{Jiang2024}. These are GCN models which also use Rough Path Theory, a mathematical framework for efficient descriptions of paths (such as joint trajectories), which has the desirable property of being invariant to time-reparametrisations, and has therefore been suggested to demonstrate enhanced robustness to variable or irregularly sampled data streams \cite{Liao2021}. For a more in-depth review of Rough Path Theory see Supplementary Section 1.
\begin{itemize}
    \item LogSigRNN (No license supplied with publication) \cite{Liao2021} is built around a \emph{LogSigRNN module}, which is a generalisation of an RNN. In this module, a GCN layer first learns the relational positions of each joint. This information is then fed into a layer that applies the log signature transform, shrinking the time dimension of the input to a fixed length whilst preserving temporal information. Finally, the outputs from this layer are fed into an appropriate RNN.
    \item GCNDevLSTM (MIT License) \cite{Jiang2024} uses a different Rough Path Theory approach called the \emph{Path Development}, which involves mapping the path using a fully trainable, parameterised linear map into the development space \cite{lou2024path}. The path development maintains the same useful properties as the signature (replacing the non-trainable log-signature layer from the LogSigRNN model), making it ideal for time-series learning whilst also being data adaptable.
\end{itemize}

\subsection{Aims and contributions}


We aim to provide an important first data degradation benchmark on the most detailed and largest 3D open dataset (the NTU-RGB+D-120 \cite{Liu2019}), and assess the robustness of five leading SHAR models to three forms of degradation that are representative of real-world issues. It is also important to note that our benchmark is not limited to the NTU-RGB+D-120 dataset, and can be further applied to other datasets where appropriate domain specific considerations apply. 


Our contributions are three-fold: 
\begin{enumerate}
\item We introduce an important first data degradation benchmark for SHAR. We demonstrate the need for this benchmark by showing that form of degradation, which has not previously been considered, has a large impact on model accuracy; at the same effective frame rate, model accuracy can vary by $>40\%$ depending on degradation type.
\item We identify that temporal regularity of frames in degraded SHAR data is likely a major driver of differences in model performance, and harness this information to improve performance of existing models by up to $>40\%$, through employing a simple inference-time mitigation approach based on interpolation.
\item We highlight how our benchmark has helped identify an important degradation-resistant SHAR model based in Rough Path Theory; the LogSigRNN SHAR model outperforms the SoTA DeGCN model in five out of six cases at low frame rates (3 FPS, common for real-world deployments), by an average accuracy of $6\%$, despite trailing the SoTA model by $11-12\%$ on un-degraded data at high frame rates (30 FPS)
\end{enumerate}

\section{Methods}
\label{sec:Method}

\subsection{NTU-RGB+D-120 dataset and degradations}
\label{subsec:data}

The NTU-RGB+D-120 dataset (license: Custom (research-only)) \cite{Liu2019} features 114,480 clips of 155 different viewpoints with 106 distinct actors for 120 different action classes. This dataset is an update of the earlier NTU-RGB+D-60 \cite{Shahroudy2016} dataset and has 48,000 more clips, 75 new viewpoints, 60 new action classes, and increased variability of target backgrounds, with the update aimed at increasing dataset variability to reduce over-fitting and provide better generalisation of trained models. Action classes include 82 daily actions (e.g. eating, writing, sitting down, moving objects), 12 health-related actions (e.g. blowing nose, vomiting, staggering, falling down), and 26 mutual actions (e.g. handshaking, pushing, hitting, hugging). Actions vary from individual full-body actions (e.g. arm swings, side kicks, jump up) to actions involving minimal body movement (e.g. OK sign, thumbs up, reading, writing) to dual person actions (e.g. carry something with other person, playing rock-paper-scissors, follow other person). 

We consider three different types of time-series degradation representing a range of forms of degradation prevalent in real-world deployments:

\begin{enumerate}
\item \textbf{Random subsampling:} We emulate random frame dropout of the time-series, where we scale the volume of randomly dropped frames from 0\% to 90\%, for direct comparison to other works and ablation studies performed in the literature.

\item \textbf{Uniform subsampling:} We emulate uniform frame dropout by retaining every $n^{th}$ frame in each sample and deleting the others. The NTU-RGB+D-120 dataset is captured at $30$ FPS, and we consider four reduced frame rates; $15$ FPS ($n$=2), $10$ FPS ($n$=3), $5$ FPS ($n$=6) and $3$ FPS ($n$=10). For relative comparison with random frame dropout these correspond to the same overall volume of deleted frames as $50\%$, $66.6\%$, $83.3\%$ and $90\%$ effective dropout rates, respectively. 

\item \textbf{Single block dropout:} We implement the erasure of single contiguous blocks of successive frames to emulate complete skeletal occlusion, skeletal segmentation failures, viewpoint obfuscation, network or onboard camera processing problems, over a large group of frames. To emulate block dropout, we randomly remove frames for a single randomly located time window in each sample. To match the same drop in FPS as those seen for uniform dropout we remove patches equivalent to $50\%$, $66.6\%$, $83.3\%$ and $90\%$ of the same length from a random initial starting position in each sample. 
\end{enumerate}

\textbf{Dropout mitigation:} We introduce a mitigation technique for frame loss where it is assumed that the positions of the dropped frames are known. The dropped frames are replaced with interpolated frames, in which the positions of all joints are estimated from linear interpolation of their positions in the adjacent available frames. In order to facilitate this interpolation we exclude the first and last frames from all of the degradation experiments with and without mitigation. This is to provide data anchors for interpolation that will not be deleted during implementation of degradation.

\subsection{SHAR models and training}
\label{subsec:models}
We consider five different SHAR models, MS-G3D Net \cite{Liu2020}, FR-HEAD \cite{Zhou2023}, DeGCN \cite{Myung2024}, LogSigRNN \cite{Liao2021} and GCN-Dev-LSTM \cite{Jiang2024}. In this study we only investigate joint-based backbones for these five models.  This is to enable fair comparison across models; only some of these models use ensembles, and not all ensembles use the same set of graph backbones. Furthermore, running ensembles is likely to be infeasible for the edge or real-time deployments that are the focus of this study.

We note that some models operate under slightly different pre-processing conditions. Following the work of \cite{Chen2021}, a large number of models in the literature, including the FR-HEAD, GCN-Dev-LSTM and DeGCN models under consideration here, adopt a pre-processing procedure of downsampling the initial samples to $64$ frames using bilinear interpolation. As this is a model-dependent decision aimed at constructing uniform batch shapes without adopting any padding techniques we perform our experiments by degrading the data streams before this step. The MS-G3D Net and LogSigRNN models do not use this pre-processing, and instead have fixed stream dimensions of $300$ frames, as this is the maximum length stream in the dataset. For streams of frames less than the maximum length the MS-G3D net model replicates each sample until the vector of length $300$ is filled, whilst for the LogSigRNN model the last frame is repeated to fill the remaining frames. We investigate and discuss the consequences of the differences in pre-processing in Section \ref{sec:Results}.

\subsection{Experimental setup}
\label{subsec:exp}

We use the cross-subject train-test split \cite{Liu2019} on the NTU-RGB+D-120 dataset to first train each of the five models, and then test them after applying the three degradations to the test data. We apply the degradation before any additional pre-processing steps that are internally implemented in each of the models. To measure the impact of the effective dropout rate on each model's performance we first obtain the baseline performance for each of the models on un-degraded data. We use pre-trained weights that are supplied with the MS-G3D Net and FR-HEAD models, and therefore the baseline performance for these models exactly matches that recorded in the corresponding papers. We train the other three models ourselves on an NVIDIA A100 PCIe GPU under the default guidance and hyper parameters reported in each model's paper; the baseline performance for these models is consistent with those previously reported. Baseline performances for all five models under consideration are presented in Table \ref{Tab:Baseline}. Inference for all 5 models are performed on an NVIDIA A100 PCIe GPU. We provide a summary of inference statistics for the models in Supplementary Section 2. We also repeat all experiments, including training and inference, for the cross-setup train-test split \cite{Liu2019} on the NTU-RGB+D-120 dataset.

\section{Results}
\label{sec:Results}

\begin{figure}[ht]
\centering
\includegraphics[width=\textwidth]{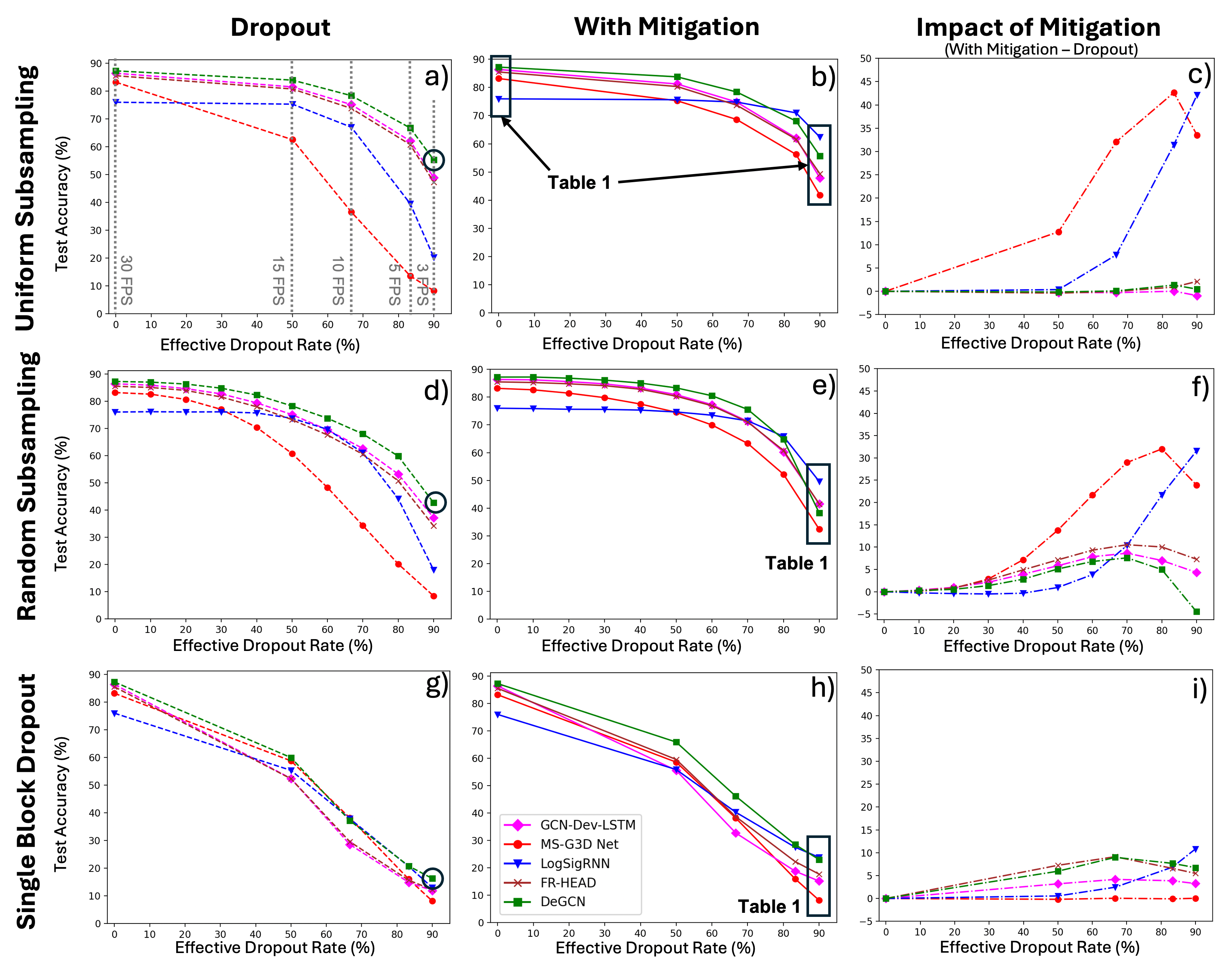}
\caption{Test accuracy versus effective dropout rate for three different experiments, Uniform Subsampling ($a, b, c$), Random Subsampling ($d, e, f$) and  Single block Dropout ($g, h, i$) and five SHAR models (see key in panel $h$). The columns show experiments where removed frames are not replaced (left), where linear interpolation is used to mitigate degradation at inference time by replacing missing frames (central), and the difference (right). Panel $a$ also shows equivalent frame-rates in FPS for translation from effective dropout rate. Black circles in $a, d, g$ highlight DeGCN performance at $90\%$ effective dropout rate across all three experiments, and black rectangles in $b, e, h$ show data points that are presented in Table \ref{Tab:Baseline}; all data are provided in Supplementary Tables 1-6.}
\label{fig:dropout_experiments_xsub}
\end{figure}

In order to evaluate performance of the five selected models on degraded data, we use accuracies on the non-degraded NTU-RGB+D-120 dataset as our `baseline'. For this non-degraded baseline, DeGCN has SoTA performance, with $87.2\%$ and $89.1\%$ accuracy for the cross subject (X-Sub) and cross setup (X-Set) train-test splits respectively (Table \ref{Tab:Baseline} columns 1-3). In comparison, the LogSigRNN model has the worst performance out of the five models on non-degraded data, with equivalent accuracies of $76.0\%$ and $77.1\%$; $11-12\%$ lower than the DeGCN values. 
Note that hereafter, we focus solely on the cross subject analysis for simplicity (Fig. \ref{fig:dropout_experiments_xsub}); the cross setup analysis (Supplementary Section 4) gives results that are qualitatively the same (compare with Supplementary Fig. 1), and therefore acts here as independent verification of our results, highlighting their robustness.

\begin{table}
\caption{Test accuracy on the NTU-RGB+D-120 dataset for each of the 5 models (1st column) recorded for both the cross subject (X-Sub) and cross setup (X-Set) train-test splits. The second and third columns correspond to baseline performances of each model. The remaining columns correspond to test accuracies recorded at $90\%$ effective dropout (equivalent to $3$ FPS) after mitigation has been applied for uniform subsampling (columns four and five), random subsampling (columns six and seven) and single block dropout (columns eight and nine). The highest accuracy models under each experiment are highlighted in bold.}
\begin{tabular}{lllllllll}
\toprule
\textbf{Dropout Rate} & \multicolumn{2}{c}{0\% (Baseline)} & \multicolumn{6}{c}{90\% (with mitigation)} \\\cmidrule(lr){1-3}\cmidrule(lr){4-9}
Sampling & \multicolumn{2}{c}{\textbf{None (30 FPS)}} & \multicolumn{2}{c}{Uniform (3 FPS)} & \multicolumn{2}{c}{Random} & \multicolumn{2}{c}{Single Block}
\\\cmidrule(lr){1-1}\cmidrule(lr){2-3}\cmidrule(lr){4-5}\cmidrule(lr){6-7}\cmidrule(lr){8-9}
\textbf{Models} & X-Sub & X-Set & X-Sub & X-Set & X-Sub & X-Set & X-Sub & X-Set \\\midrule
MS-G3D Net & 83.2\% & 84.2\% & 41.8\% & 50.4\% & 32.4\% & 42.8\% & 8.2\% & 7.6\%  \\
FR-HEAD & 85.5\% & 87.3\%  & 49.4\% & 49.3\% & 41.5\% & 42.0\% & 17.7\% & 16.4\% \\ 
DeGCN  & \textbf{87.2\%} & \textbf{89.1\%} & 55.7\% & 51.9\% & 38.3\% & 40.4\% & 23.0\% & \textbf{23.7\%} \\ 
LogSigRNN & 76.0\% & 77.1\% & \textbf{62.4\%} & \textbf{63.3\%} & \textbf{49.6\%} & \textbf{52.1\%} & \textbf{23.9\%} & 20.8\%  \\
GCNDevLSTM  & 86.4\% & 88.1\%  & 47.9\% & 46.0\% & 41.5\%  & 37.5\%& 15.2\% & 13.0\%  \\ 
\bottomrule
\end{tabular}
\label{Tab:Baseline}
\end{table}


\noindent{\textbf{Characterising and understanding the impact of degradation.}} As effective dropout rate increases, all five models show progressively decreasing performance. This occurs across all three types of data degradation (uniform subsampling, random subsampling and single block dropout; Fig. \ref{fig:dropout_experiments_xsub} a,d,g respectively). However, not all models exhibit the same behaviour in terms of how performance changes with increasing dropout rate. This is illustrated by the increasing range of model performance as we move to higher dropout rates; at baseline ($0\%$ dropout) the range of model performance across the five different models is $11\%$, whilst the range of model performance is significantly higher at $90\%$ effective dropout rate ($47\%$ for uniform subsampling and $34\%$ for random subsampling; Fig. \ref{fig:dropout_experiments_xsub} a,d). 
It is also notable that the DeGCN, FR-HEAD and GCN-Dev-LSTM models, which all employ the same resampling pre-processing step (detailed in Section \ref{subsec:models}), show very similar sensitivity to degradation (similarity of green, pink and brown curves in Fig. \ref{fig:dropout_experiments_xsub} a,d). Overall the DeGCN model maintains SoTA (albeit decreased) performance for all dropout rates, and the similarity of the FR-HEAD and GCN-Dev-LSTM performance curves to this model's suggests that resampling data-streams to a fixed length during pre-processing is effective for mitigating the impact of degraded time-series. But the behaviour of these three models is markedly dissimilar to that of the two models without this pre-processing (MS-G3D Net and LogSigRNN, red and blue curves respectively). The consistently worst model at the highest effective dropout rate across all three degradation methods is MS-G3D Net, but the LogSigRNN model exhibits the most striking behaviour; there is no significant drop in model accuracy ($<1\%$) for effective dropout rates up to $40-50\%$ for uniform and random subsampling. This highlights the effectiveness of Rough Path Theory based models to be robust to missing data, and we return to this topic at the end of this section.


However, few of these findings apply to the results for single block dropout; there is no clear distinction between behaviour of different models with increasing dropout rate, and no increase in the range of model performance at high dropout rates in comparison to baseline (i.e. all model lines have similar profiles in Fig. \ref{fig:dropout_experiments_xsub} g). 
This highlights a key finding; that the form or type of degradation can have a major impact on model performance. This is illustrated by DeGCN's performance at $90\%$ effective dropout rate; $55\%$, $42\%$ and $16\%$ for uniform subsampling, random subsampling and single block dropout respectively (see black circled points in Fig. \ref{fig:dropout_experiments_xsub} a,d,g). For the same percentage of removed frames, the performance of the SoTA model can vary by almost $40\%$ depending on which subset of frames are removed. We suggest that it is variation in the regularity of the residual frames after dropout that drives this major difference; as the frequency of sampling becomes more irregular (from uniform subsampling to random subsampling to single block dropout) the performance of each model reduces (compare Fig. \ref{fig:dropout_experiments_xsub} a,d,g). A simple explanation is that increased grouping of removed frames increases the likelihood that clustered discriminative frames from an action sequence are removed; this contributes to a consistent drop in performance across models and one which cannot be mitigated by model architectural advantages. 
It is also important to note that uniform subsampling and single block dropout degradation methods appear to represent end-members that bound the performance of random subsampling. This therefore justifies the need for our new degradation benchmark; ablation studies that only employ random subsampling are not sufficient, as this form of degradation alone does not capture bounds on real-world performance that would result from reductions in frame-rate (uniform sampling) or total occlusion of the subject for a portion of the action (block dropout).

\noindent{\textbf{Utility of benchmark: simple mitigation to degraded data}}. Our new benchmark has helped show that 1) bilinear interpolation, when used as a pre-processing step during training and inference, appears to increase model resilience to degraded SHAR data, and 2) regularity of degraded input has a strong impact on model performance. We now use these insights to suggest a simple post-hoc mitigation technique that could be widely applied at inference time, even for models which do not include this pre-processing. We rerun all experiments but replace missing frames with linearly interpolated frames as described in Section \ref{subsec:data}. This new analysis (Fig. \ref{fig:dropout_experiments_xsub} b, e, h) highlights three key results. First, the models with the resampling pre-processing step (FR-HEAD, GCN-Dev-LSTM \& DeGCN) have a negligible mean performance improvement of $0.8\%$ when the additional mitigation is applied. Second, MS-G3D Net shows a large increase in performance ($>40\%$ for $80\%$ effective dropout rate in Fig. \ref{fig:dropout_experiments_xsub} c) and now shows similar behaviour with increasing dropout as these three models (contrast red curves in Fig. \ref{fig:dropout_experiments_xsub} b, e, with a, d). Together, these two results suggest that this post-hoc interpolation is fulfilling a similar function as the pre-processing, with similar end effect. Finally, the LogSigRNN model also shows a very large performance improvement, especially at the highest dropout rates, where it now consistently outperforms the DeGCN model to become SoTA.

\begin{figure}[ht]
\centering
\includegraphics[width=0.8\textwidth]{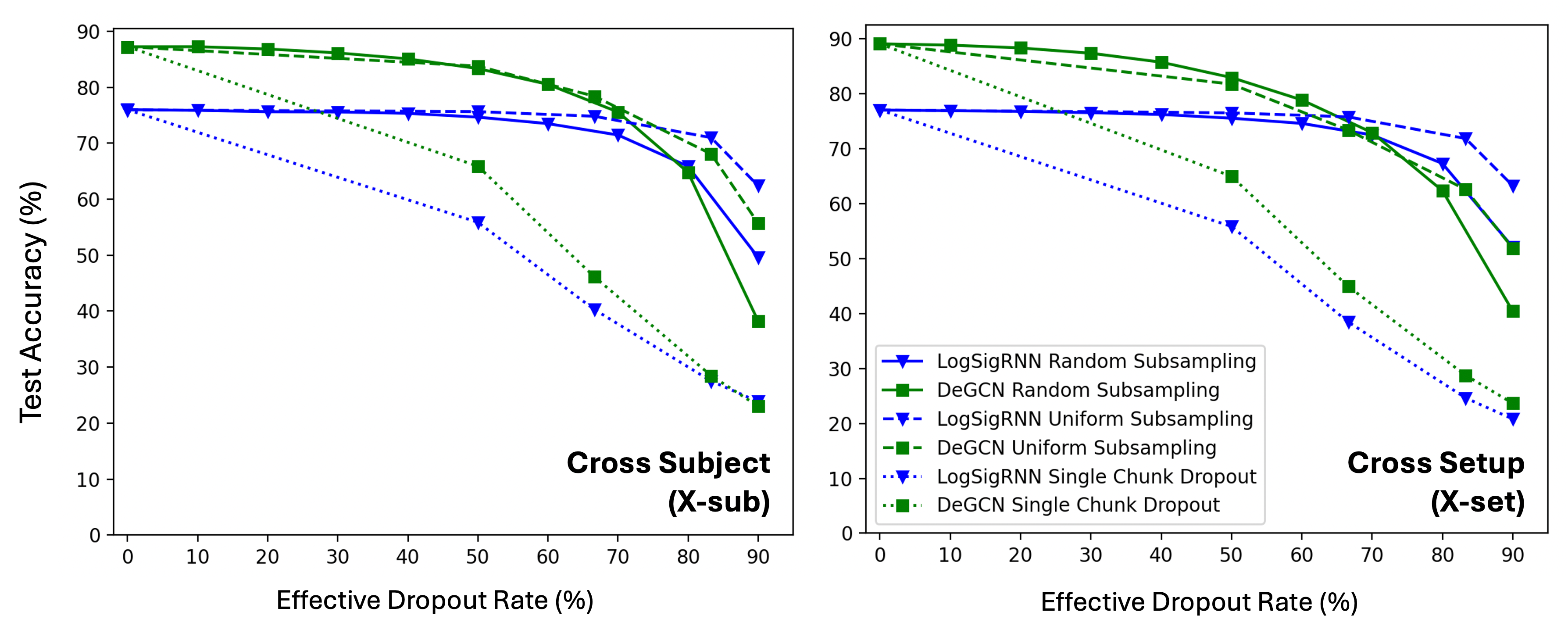}
\caption{Comparison of LogSigRNN (blue triangles) and DeGCN (green squares) models with mitigation through post-hoc linear interpolation, for random subsampling (dashed lines), uniform subsampling (solid lines) and single block dropout (dotted lines). Results are shown for both cross subject (left) and cross setup (right) train-test splits.}
\label{fig:logsig_vs_degcn}
\end{figure}

\noindent{\textbf{Utility of benchmark: identification of degradation-resilient model architecture}}. This result highlights how our new benchmark has helped identify an important degradation-resilient SHAR model. The LogSigRNN model achieves SoTA performance for effective dropout rates of $>80\%$ (i.e.~$<5$ FPS) for uniform subsampling and random subsampling, beating the DeGCN model by $6.7\%$ and $11.3\%$ at $90\%$ effective dropout rate ($=3$ FPS). This is particularly impressive when we note that the DeGCN model has $11\%$ higher accuracy than the LogSigRNN model on un-degraded data (Table \ref{Tab:Baseline}); together this represents a relative increase in performance of $19\% - 23\%$ of the LogSigRNN model with respect to DeGCN, as the effective dropout rate increases from 0\% to 90\%. As the highest dropout rates ($>80\%$) represent common real-world frame rates caused by either different capture equipment or network latency, the ascension of a different model architecture over the DeGCN model is a particularly important result.
In addition to the improved performance of the LogSigRNN model at the highest dropout rates, the LogSigRNN model also now maintains its baseline performance down to $10$ FPS ($66.7\%$ effective dropout rate) for the uniform subsampling and down to to $60\%$ effective dropout rate for the random subsampling, with a performance drops of $<1.5\%$ in each case (see blue curves in Fig. \ref{fig:dropout_experiments_xsub} b, e). All results are qualitatively consistent across both the cross subject and cross setup train-test splits (Fig. \ref{fig:logsig_vs_degcn}), which supports the robustness of these findings.




\section{Discussion}

Our new benchmark will support the community’s ability to assess, improve and develop new SHAR models and ensure their suitability for practical real-world deployment. Furthermore, we have identified a model from the Rough Path Theory (RPT) domain which has enhanced performance for SHAR versus the SoTA model at low frame rates. This is particularly relevant for deployment of SHAR models on edge networks, where recent studies have shown that frame rates can be as low as 3 FPS due to excessive dropout and fluctuating latency \cite{Wang2024}.

These findings warrant further investigation of the utility of RPT models in degraded data environments. Future research should aim to combine the robustness to data degradation exhibited by the LogSigRNN model with the higher baseline performance of more recent graph-based models such as DeGCN. This could be achieved by integrating the signature transform layer into the DeGCN architecture. The utility of broader RPT-related architectures could also be investigated, such as Neural Controlled Differential Equations (CDE) \cite{kidger2020neuralcde}, which have been shown to outperform RNN-based models on time-series classification tasks with missing data and have previously been combined with RPT signatures \cite{morrill2021a}. However, we also note that not all RPT models may have the same tolerance of degraded time-series exhibited by the LogSigRNN model; even within this study the GCN-Dev-LSTM model, which employs a very different form of RPT in its architecture, does not show the same beneficial features (contrast pink and blue curves in Fig. \ref{fig:dropout_experiments_xsub} b, e).



We also propose that future work should focus on augmenting training datasets with varying simulated degradations, in order to improve model robustness to these variations. This could involve both situations where the statistical structure of degradation is known, and where this is unknown. A potential approach could be adapted from work on wireless channel signal-to-noise-ratio (SNR) degradation \cite{10066513}, where a degradation-aware model was developed that uses attention modules to feed estimated degradation statistics into the model and influence the model's decision process.

This work focuses on benchmarking model performance against forms of degradation that involve total loss of all information in one or more frames; these can be considered as first-order errors. In future we suggest the benchmark to be extended to second-order errors, which would include spatial rather than temporal degradation of the action sequences (partial occlusion \cite{Kostis2022}, failed skeletal segmentation \cite{Riou2024}) or epistemic uncertainties e.g. from jitter introduced by imperfect pose-estimation on RGB video streams \cite{zeng2022smoothnet}. 

In this study, we have demonstrated the importance and value of benchmarking computer vision models against degraded scenarios that represent the limitations of real-world deployments. Our systematic assessment of a range of SHAR models on degraded data from the NTU-RGB+D-120 dataset has shown that the form and structure of degradation has a significant influence on the degraded performance of the models. We have demonstrated the value of this benchmark by leveraging our insights to suggest a new simple and practical mitigation that can be applied at inference, and to identify a particularly degradation-resistant model that overtakes the existing state-of-the-art model at low, real-world frame rates. We hope that broader adoption of our systematic approach to assessing model performance on purposely-degraded video datasets will foster the development of a new generation of degradation-resilient models across a wide variety of computer vision domains.

\noindent{\bf {Acknowledgement:}} Research was sponsored by the Army and was accomplished under Cooperative Agreement Number W911NF-22-2-0161. The views and conclusions contained in this document are those of the authors and should not be interpreted as representing the official policies, either expressed or implied, of the U.S. Army or the U.S. Government. The U.S. Government is authorized to reproduce and distribute reprints for Government purposes notwithstanding any copyright notation herein.

Some of the computations described in this research were performed using the Baskerville Tier 2 HPC service (\url{https://www.baskerville.ac.uk/}). Baskerville was funded by the EPSRC and UKRI through the World Class Labs scheme (EP/T022221/1) and the Digital Research Infrastructure programme (EP/W032244/1) and is operated by Advanced Research Computing at the University of Birmingham.

{
    \small
    \bibliographystyle{ieeetr}
    \bibliography{HAR_bib}
}

\newpage
\setcounter{page}{1}
\setcounter{section}{0}

\begin{center}
\textbf{\large Supplementary Material}
\end{center}

\section{Review of Rough Path Theory for Skeletal Human Action Recognition}

\noindent{\bf{SHAR methods that use Rough Path Theory:}} Rough Path Theory provides a mathematical framework for analysing complex, irregular signals by capturing the signal's geometric structure beyond traditional Euclidean methods \cite{Lyons2007}. The response to these irregular signals is a \emph{rough} path. Using Euclidean calculus, a path must be sufficiently differentiable (smooth) in order for integration between data points. This condition fails where paths are sufficiently noisy, such as when considering the trajectory of skeletal joints. 

Rough paths circumvent this problem by augmenting the path with a second-order process which lifts the integration of a path, using iterated integrals, into a higher dimensional space. The basis of this higher dimensional space contains different combinations of the dimensions of variation of the path and hence combines both measures over the interval (Euclidean calculus) and cross terms (non-Euclidean calculus) measuring correlations between variables. By lifting into successively higher dimensions, more complex features describing the relationship between joints (cross terms) can be encoded, such as relative joint velocities and rotational motion. The collection of iterated integrals of increasing dimensions over the path is called the \emph{signature}. For efficient machine learning, higher dimensions are often problematic for practical operation and deployment, but crucially the signature is much smaller than the raw signals. 

The signature is equipped with three important properties that we can leverage for SHAR:

\begin{itemize}
\item{{\bf Hierarchical representation:} The signature is broken into a series of levels, with each level containing increasingly higher order iterated integrals. Lower order terms describe global dynamics whilst higher order terms describe successively more fine-grained dynamics. This presents a graded feature set which summarises the path. For practical machine learning purposes this feature set can be truncated to compute a finite length signature of the path, known as the \emph{truncated signature}. This is a minimal and low-dimensional representation, desirable for machine learning.}
\item{{\bf Natural robustness to noise:} Due to the iterated integral-based formulation, iterated integrals act like a low-pass filter. Joint noise (jitter) has a mean of zero and is uncorrelated in time, so when integrating over time-intervals such contributions systematically vanishes.} 
\item{{\bf Invariance to reparametrisation:} The signature of the path space only depends on the order of events and measures the geometry of the path shape. This disregards speed variations and hence reparametrisation does not affect the value of the integral. This property makes the signature consistent across variable execution rates and robust to variable sampling rates and missing frames.}
\end{itemize}

Due to these three properties, this approach enables the modelling of temporal dependencies and kinematic relationships in a compact, interpretable manner, making it well-suited for action classification, detection, and anomaly tasks. For in-depth reviews of the use of rough path theory for machine learning please see references \cite{Lyons2014, Chevyrev2016} and references therein. 

The first rough path theory models for human action recognition focused on using rough path theory as a compact feature extractor for neighbouring joints applied to hand gesture recognition \cite{Li2017} and SHAR \cite{Li2019}. Later formulations focused on introducing angles between joints \cite{Ahmad2019} and to non-local joint pairs and triplets \cite{Yang2022}. More recently, with the up-take in GCN methods in this field, hand crafted features have become obsolete. Instead, two new methods that combine GCNs and rough path theory with RNNs and LSTMs show promising results on SHAR tasks; LogSigRNN and GCN-Dev-LSTM.

\noindent{\bf{LogSigRNN:}} GCNs provide an optimal solution for extracting the spatial features for skeletal data. However, they have no effect on the length of the data sequence, and when data sequences have large and variable length, RNNs, such as LSTMs, suffer from extraneous training costs. To manage computational costs, data streams are often padded or downsampled/resampled. To maintain the shape of the length of the sequence the authors of LogSigRNN \cite{Liao2021} feed the logarithm of the truncated signature into the RNN as the signature dimensions are constant and a graded summary of the path is sufficient for RNN learning. This compact representation of the feature set considerably reduces the size of GCN-LSTM models, speeding up computation times. However, this does come at a cost; despite not having any learnable parameters, the signature is non-differentiable over long sequences due to numerical instabilities. To overcome this limitation the log-signature is applied to overlapping time windows, creating local descriptors of the skeletal time-series.  

In the LogSigRNN model \cite{Liao2021}, the raw coordinates of each joint as a function of time are fed into the network. These are passed through an embedding or graph convolution network layer to learn relational positions of each joint, referred to as a path transformation layer. The remainder of the network is a generalisation of an RNN, which the authors call a LogSigRNN module. The embedding in path space computed by the path transformation layer is first fed into a layer which applies the log signature transform. This layer does not have any trainable parameters, but shrinks the time dimension of the input stream whilst preserving the temporal information. Each representation is now of the same fixed length, and these representations are then fed into an appropriate RNN.

\noindent{\bf{GCNDevLSTM:}} Without learnable parameters, signature based methods have to be computed to increasingly higher dimensions to capture more information. Due to the curse of dimensionality, over-fitting rapidly becomes a problem with increasing numbers of signature terms. To avoid this problem and to increase the data-adaptability of rough path theory models, the authors of \cite{Jiang2024} consider a different rough path theory approach called the \emph{Path Development}. Exploiting the representation of the matrix Lie group, the path is mapped using a fully trainable, parametrised linear map into the development space \cite{lou2024path}. The path development maintains the same useful properties as the signature, making it ideal for time-series learning. This layer plays the same role as the log-signature layer in the LogSigRNN model, with the added benefit of being data adaptable. However, the set of additional trainable parameters does lead to an increase in model size. 

\section{Inference GPU resource usage}

Our benchmark evaluates test performance of five Skeletal Human Action Recognition models on systematically degraded data. We train and test each model using their default parameters reported in their respective publications, as described in the main text. As the MS-G3D Net and FR-HEAD models contain pre-trained weights we do not need to perform any model retraining or fine-tuning and therefore do not report training times for these models. Test times and training times for the remaining models are reported in Table \ref{Tab:gpu}. All models are trained and tested on an NVIDIA A100 80 GB Tensor Core PCIe GPU with 3rd Gen AMD EPYC Milan processors. We note that test time includes the time to perform degradation over the whole dataset and to test the whole dataset. 

\begin{table}[h!]
\caption{Training and test times for the 5 models used in the main paper.}
\begin{tabular}{llll}
\toprule
Models & Training time & Test time  & \multicolumn{1}{c}{Notes} \\\midrule
MS-G3D Net   & N/A & 7-8 mins & Pre-trained weights supplied \\
FR-HEAD      & N/A & 10-11 mins & Pre-trained weights supplied \\
DeGCN        & 11 hrs & 1-2 mins & \\
LogSigRNN    & 23 hrs & 4-5 mins & \\
GCN-Dev-LSTM & 48 hrs & 10-11 mins & \\
\bottomrule
\end{tabular}
\label{Tab:gpu}
\end{table}




\section{Data Tables}

For reference and completeness, we present the data tables for all data plotted on Figure 1 from the main paper (Tables \ref{Tab:x_sub_uniform_without} - \ref{Tab:x_sub_single_with}) and for Supplementary Figure \ref{fig:dropout_experiements_xset} (Tables \ref{Tab:x_setup_uniform_without} - \ref{Tab:x_setup_single_with}). 

\subsection{Cross-subject}

\begin{table}[h!]
\caption{Data table reporting model accuracies for the cross-subject train-test split for the uniform subsampling experiment without mitigation.}
\begin{tabular}{llllll}
\toprule
 & \multicolumn{5}{c}{Models} \\\cmidrule(lr){2-6}
Frame rate & MS-G3D Net & FR-HEAD & DeGCN & LogSigRNN & GCNDevLSTM \\\midrule
$30$ FPS ($0\%$) & $83.2\%$ & $85.5\%$ & $87.2\%$ & $76.0\%$ & $86.4\%$ \\ 
$15$ FPS ($50\%$)   & $62.6\%$ & $80.7\%$ & $83.9\%$ & $75.3\%$ & $60.9\%$ \\
$10$ FPS ($66.6\%$) & $36.6\%$ & $73.8\%$ & $78.3\%$ & $67.0\%$ & $75.1\%$ \\
$5$ FPS ($83.3\%$)  & $13.6\%$ & $60.8\%$ & $66.7\%$ & $39.6\%$ & $62.1\%$ \\
$3$ FPS ($90\%$)    & $8.3\%$ & $20.3\%$ & $55.3\%$ & $20.3\%$ & $49.5\%$ \\
\bottomrule
\end{tabular}
\label{Tab:x_sub_uniform_without}
\end{table}

\begin{table}[h!]
\caption{Data table reporting model accuracies for the cross-subject train-test split for the uniform subsampling experiment with mitigation.}
\begin{tabular}{llllll}
\toprule
 & \multicolumn{5}{c}{Models} \\\cmidrule(lr){2-6}
Frame rate & MS-G3D Net & FR-HEAD & DeGCN & LogSigRNN & GCNDevLSTM \\\midrule
$30$ FPS ($0\%$) & $83.2\%$ & $85.5\%$ & $87.2\%$ & $76.0\%$ & $86.4\%$ \\ 
$15$ FPS ($50\%$)   & $75.3\%$ & $80.3\%$ & $83.8\%$ & $75.6\%$ & $81.2\%$ \\
$10$ FPS ($66.6\%$) & $68.6\%$ & $73.7\%$ & $78.4\%$ & $74.8\%$ & $74.8\%$ \\
$5$ FPS ($83.3\%$)  & $56.3\%$ & $61.7\%$ & $68.1\%$ & $71.0\%$ & $62.1\%$ \\
$3$ FPS ($90\%$)    & $41.8\%$ & $49.4\%$ & $55.7\%$ & $62.4\%$ & $47.9\%$ \\
\bottomrule
\end{tabular}
\label{Tab:x_sub_uniform_with}
\end{table}

\begin{table}[h!]
\caption{Data table reporting model accuracies for the cross-subject train-test split for the random subsampling experiment without mitigation.}
\begin{tabular}{llllll}
\toprule
 & \multicolumn{5}{c}{Models} \\\cmidrule(lr){2-6}
Dropout Rate ($\%$) & MS-G3D Net & FR-HEAD & DeGCN & LogSigRNN & GCNDevLSTM \\\midrule
$0\%$ (Baseline) & $83.2\%$ & $85.5\%$ & $87.2\%$ & $76.0\%$ & $86.4\%$ \\ 
$10\%$ & $82.6\%$ & $85.0\%$ & $87.0\%$ & $76.1\%$ & $85.6\%$ \\
$20\%$ & $80.6\%$ & $84.0\%$ & $86.3\%$ & $76.0\%$ & $84.6\%$ \\
$30\%$ & $77.0\%$ & $81.5\%$ & $84.8\%$ & $76.0\%$ & $82.7\%$ \\
$40\%$ & $70.4\%$ & $77.9\%$ & $82.3\%$ & $75.6\%$ & $79.4\%$ \\
$50\%$ & $60.8\%$ & $73.2\%$ & $78.2\%$ & $73.7\%$ & $75.1\%$ \\
$60\%$ & $48.3\%$ & $67.6\%$ & $73.7\%$ & $69.7\%$ & $69.4\%$ \\
$70\%$ & $34.3\%$ & $60.5\%$ & $68.0\%$ & $61.1\%$ & $62.6\%$ \\
$80\%$ & $20.2\%$ & $50.7\%$ & $59.8\%$ & $44.2\%$ & $50.7\%$ \\
$90\%$ &  $8.5\%$ & $34.2\%$ & $42.7\%$ & $18.1\%$ & $34.2\%$ \\
\bottomrule
\end{tabular}
\label{Tab:x_sub_random_without}
\end{table}

\begin{table}[h!]
\caption{Data table reporting model accuracies for the cross-subject train-test split for the random subsampling experiment with mitigation.}
\begin{tabular}{llllll}
\toprule
 & \multicolumn{5}{c}{Models} \\\cmidrule(lr){2-6}
Dropout Rate ($\%$) & MS-G3D Net & FR-HEAD & DeGCN & LogSigRNN & GCNDevLSTM \\\midrule
$0\%$ (Baseline) & $83.2\%$ & $85.5\%$ & $87.2\%$ & $76.0\%$ & $86.4\%$ \\ 
$10\%$ & $82.6\%$ & $85.3\%$ & $87.2\%$ & $75.9\%$ & $86.2\%$ \\
$20\%$ & $81.4\%$ & $84.8\%$ & $86.8\%$ & $75.6\%$ & $85.6\%$ \\
$30\%$ & $79.8\%$ & $84.1\%$ & $86.1\%$ & $75.6\%$ & $84.8\%$ \\
$40\%$ & $77.5\%$ & $82.8\%$ & $85.1\%$ & $75.3\%$ & $83.3\%$ \\
$50\%$ & $74.4\%$ & $80.3\%$ & $83.3\%$ & $74.7\%$ & $80.9\%$ \\
$60\%$ & $70.0\%$ & $76.9\%$ & $80.5\%$ & $73.5\%$ & $77.3\%$ \\
$70\%$ & $63.3\%$ & $71.0\%$ & $75.6\%$ & $71.5\%$ & $71.3\%$ \\
$80\%$ & $52.2\%$ & $60.7\%$ & $64.8\%$ & $65.9\%$ & $60.2\%$ \\
$90\%$ &  $32.4\%$ & $41.5\%$ & $38.3\%$ & $49.6\%$ & $41.5\%$ \\
\bottomrule
\end{tabular}
\label{Tab:x_sub_random_with}
\end{table}

\begin{table}[h!]
\caption{Data table reporting model accuracies for the cross-subject train-test split for the single block dropout experiment without mitigation.}
\begin{tabular}{llllll}
\toprule
 & \multicolumn{5}{c}{Models} \\\cmidrule(lr){2-6}
Frame rate & MS-G3D Net & FR-HEAD & DeGCN & LogSigRNN & GCNDevLSTM \\\midrule
$30$ FPS ($0\%$) & $83.2\%$ & $85.5\%$ & $87.2\%$ & $76.0\%$ & $86.4\%$ \\ 
$15$ FPS ($50\%$)   & $58.8\%$ & $52.3\%$ & $59.9\%$ & $55.3\%$ & $52.4\%$ \\
$10$ FPS ($66.6\%$) & $38.1\%$ & $28.6\%$ & $37.2\%$ & $37.9\%$ & $28.6\%$ \\
$5$ FPS ($83.3\%$)  & $16.1\%$ & $14.9\%$ & $20.8\%$ & $20.7\%$ & $14.9\%$ \\
$3$ FPS ($90\%$)    & $8.2\%$ & $11.9\%$ & $16.3\%$ & $13.0\%$ & $11.9\%$ \\
\bottomrule
\end{tabular}
\label{Tab:x_sub_single_without}
\end{table}

\begin{table}[h!]
\caption{Data table reporting model accuracies for the cross-subject train-test split for the single block dropout experiment with mitigation.}
\begin{tabular}{llllll}
\toprule
 & \multicolumn{5}{c}{Models} \\\cmidrule(lr){2-6}
Frame rate & MS-G3D Net & FR-HEAD & DeGCN & LogSigRNN & GCNDevLSTM \\\midrule
$30$ FPS ($0\%$) & $83.2\%$ & $85.5\%$ & $87.2\%$ & $76.0\%$ & $86.4\%$ \\ 
$15$ FPS ($50\%$)   & $58.6\%$ & $59.5\%$ & $65.9\%$ & $55.9\%$ & $55.6\%$ \\
$10$ FPS ($66.6\%$) & $38.2\%$ & $38.7\%$ & $46.2\%$ & $40.3\%$ & $32.8\%$ \\
$5$ FPS ($83.3\%$)  & $16.0\%$ & $22.3\%$ & $28.5\%$ & $27.5\%$ & $18.8\%$ \\
$3$ FPS ($90\%$)    & $8.2\%$ & $17.7\%$ & $23.9\%$ & $23.9\%$ & $15.2\%$ \\
\bottomrule
\end{tabular}
\label{Tab:x_sub_single_with}
\end{table}

\newpage

\subsection{Cross-setup}

\begin{table}[h!]
\caption{Data table reporting model accuracies for the cross-setup train-test split for the uniform subsampling experiment without mitigation.}
\begin{tabular}{llllll}
\toprule
 & \multicolumn{5}{c}{Models} \\\cmidrule(lr){2-6}
Frame rate & MS-G3D Net & FR-HEAD & DeGCN & LogSigRNN & GCNDevLSTM \\\midrule
$30$ FPS ($0\%$) & $84.2\%$ & $87.3\%$ & $89.1\%$ & $77.1\%$ & $88.1\%$ \\ 
$15$ FPS ($50\%$)   & $57.3\%$ & $80.6\%$ & $81.7\%$ & $74.91\%$ & $81.7\%$ \\
$10$ FPS ($66.6\%$) & $31.3\%$ & $72.3\%$ & $72.5\%$ & $64.6\%$ & $73.9\%$ \\
$5$ FPS ($83.3\%$)  & $11.5\%$ & $59.6\%$ & $60.0\%$ & $36.4\%$ & $59.8\%$ \\
$3$ FPS ($90\%$)    & $6.9\%$ & $46.8\%$ & $47.9\%$ & $19.2\%$ & $45.5\%$ \\
\bottomrule
\end{tabular}
\label{Tab:x_setup_uniform_without}
\end{table}

\begin{table}[h!]
\caption{Data table reporting model accuracies for the cross-setup train-test split for the uniform subsampling experiment with mitigation.}
\begin{tabular}{llllll}
\toprule
 & \multicolumn{5}{c}{Models} \\\cmidrule(lr){2-6}
Frame rate & MS-G3D Net & FR-HEAD & DeGCN & LogSigRNN & GCNDevLSTM \\\midrule
$30$ FPS ($0\%$) & $84.2\%$ & $87.3\%$ & $89.1\%$ & $77.1\%$ & $88.1\%$ \\ 
$15$ FPS ($50\%$)   & $78.2\%$ & $80.5\%$ & $81.8\%$ & $76.5\%$ & $81.6\%$ \\
$10$ FPS ($66.6\%$) & $71.9\%$ & $72.8\%$ & $73.4\%$ & $75.8\%$ & $74.1\%$ \\
$5$ FPS ($83.3\%$)  & $61.7\%$ & $61.3\%$ & $62.6\%$ & $71.9\%$ & $60.4\%$ \\
$3$ FPS ($90\%$)    & $50.4\%$ & $49.3\%$ & $51.9\%$ & $63.3\%$ & $46.0\%$ \\
\bottomrule
\end{tabular}
\label{Tab:x_setup_uniform_with}
\end{table}

\begin{table}[h!]
\caption{Data table reporting model accuracies for the cross-setup train-test split for the random subsampling experiment without mitigation.}
\begin{tabular}{llllll}
\toprule
 & \multicolumn{5}{c}{Models} \\\cmidrule(lr){2-6}
Dropout Rate ($\%$) & MS-G3D Net & FR-HEAD & DeGCN & LogSigRNN & GCN-Dev-LST \\\midrule
$0\%$ (Baseline) & $84.2\%$ & $87.3\%$ & $89.1\%$ & $77.1\%$ & $88.1\%$ \\ 
$10\%$ & $83.2\%$ & $86.1\%$ & $88.3\%$ & $76.9\%$ & $87.3\%$ \\
$20\%$ & $80.5\%$ & $83.8\%$ & $86.7\%$ & $76.9\%$ & $85.6\%$ \\
$30\%$ & $75.7\%$ & $80.3\%$ & $84.0\%$ & $76.5\%$ & $82.9\%$ \\
$40\%$ & $67.7\%$ & $75.3\%$ & $79.8\%$ & $75.7\%$ & $78.7\%$ \\
$50\%$ & $55.8\%$ & $69.3\%$ & $73.6\%$ & $73.3\%$ & $73.2\%$ \\
$60\%$ & $41.9\%$ & $63.2\%$ & $66.6\%$ & $68.0\%$ & $66.6\%$ \\
$70\%$ & $27.0\%$ & $57.3\%$ & $59.7\%$ & $57.8\%$ & $59.2\%$ \\
$80\%$ & $15.2\%$ & $49.5\%$ & $51.5\%$ & $40.4\%$ & $49.9\%$ \\
$90\%$ &  $7.1\%$ & $35.1\%$ & $37.7\%$ & $16.8\%$ & $33.7\%$ \\
\bottomrule
\end{tabular}
\label{Tab:x_setup_random_without}
\end{table}

\begin{table}[h!]
\caption{Data table reporting model accuracies for the cross-setup train-test split for the random subsampling experiment with mitigation.}
\begin{tabular}{llllll}
\toprule
 & \multicolumn{5}{c}{Models} \\\cmidrule(lr){2-6}
Dropout Rate ($\%$) & MS-G3D Net & FR-HEAD & DeGCN & LogSigRNN & GCNDevLSTM \\\midrule
$0\%$ (Baseline) & $84.2\%$ & $87.3\%$ & $89.1\%$ & $77.1\%$ & $88.1\%$ \\ 
$10\%$ & $83.9\%$ & $87.0\%$ & $88.9\%$ & $76.9\%$ & $88.04\%$ \\
$20\%$ & $83.1\%$ & $86.5\%$ & $88.4\%$ & $76.8\%$ & $87.3\%$ \\
$30\%$ & $81.9\%$ & $85.6\%$ & $87.4\%$ & $76.6\%$ & $86.50\%$ \\
$40\%$ & $80.3\%$ & $84.2\%$ & $85.8\%$ & $76.3\%$ & $84.7\%$ \\
$50\%$ & $77.8\%$ & $81.7\%$ & $83.0\%$ & $75.6\%$ & $81.8\%$ \\
$60\%$ & $74.8\%$ & $77.8\%$ & $78.9\%$ & $74.6\%$ & $77.5\%$ \\
$70\%$ & $70.1\%$ & $71.7\%$ & $73.0\%$ & $72.5\%$ & $70.6\%$ \\
$80\%$ & $61.7\%$ & $61.5\%$ & $62.4\%$ & $67.3\%$ & $59.3\%$ \\
$90\%$ &  $42.8\%$ & $42.0\%$ & $40.4\%$ & $52.1\%$ & $37.5\%$ \\
\bottomrule
\end{tabular}
\label{Tab:x_setup_random_with}
\end{table}

\begin{table}[h!]
\caption{Data table reporting model accuracies for the cross-setup train-test split for the single block dropout experiment without mitigation.}
\begin{tabular}{llllll}
\toprule
 & \multicolumn{5}{c}{Models} \\\cmidrule(lr){2-6}
Frame rate & MS-G3D Net & FR-HEAD & DeGCN & LogSigRNN & GCNDevLSTM \\\midrule
$30$ FPS ($0\%$)                & $84.2\%$ & $87.3\%$ & $89.1\%$ & $77.1\%$ & $88.1\%$ \\ 
$15$ FPS ($50\%$)               & $53.8\%$ & $49.5\%$ & $55.2\%$ & $52.5\%$ & $49.8\%$ \\
$10$ FPS ($66.3\%$)             & $31.2\%$ & $26.2\%$ & $30.5\%$ & $34.2\%$ & $25.3\%$ \\
$5$ FPS ($83.3\%$)              & $13.3\%$ & $13.9\%$ & $16.3\%$ & $18.2\%$ & $11.7\%$ \\
$3$ FPS ($90\%$)                & $7.7\%$ & $11.5\%$ & $14.2\%$ & $11.7\%$ & $10.3\%$ \\
\bottomrule
\end{tabular}
\label{Tab:x_setup_single_without}
\end{table}

\begin{table}[h!]
\caption{Data table reporting model accuracies for the cross-setup train-test split for the single block dropout experiment with mitigation.}
\begin{tabular}{llllll}
\toprule
 & \multicolumn{5}{c}{Models} \\\cmidrule(lr){2-6}
Frame rate & MS-G3D Net & FR-HEAD & DeGCN & LogSigRNN & GCNDevLSTM \\\midrule
$30$ FPS ($0\%$)                & $84.2\%$ & $87.3\%$ & $89.1\%$ & $77.1\%$ & $88.1\%$ \\ 
$15$ FPS ($50\%$)               & $53.9\%$ & $57.7\%$ & $65.0\%$ & $55.9\%$ & $56.0\%$ \\
$10$ FPS ($66.3\%$)             & $31.9\%$ & $36.7\%$ & $44.9\%$ & $38.5\%$ & $39.9\%$ \\
$5$ FPS ($83.3\%$)              & $13.3\%$ & $20.5\%$ & $28.8\%$ & $24.6\%$ & $16.9\%$ \\
$3$ FPS ($90\%$)                & $7.6\%$ & $16.4\%$ & $23.7\%$ & $20.8\%$ & $13.0\%$ \\
\bottomrule
\end{tabular}
\label{Tab:x_setup_single_with}
\end{table}
\newpage
\section{Cross-setup Analysis}

As reported in the main paper we repeat all experiments on the cross-setup train-test split. The qualitative results from the cross-setup train-test split in almost all cases are identical to the cross-subject train-test split that are described in Section 3 of the main paper. This section is dedicated to highlighting the minor differences in results between the different train-test splits. 

\begin{figure}[h!]
\centering
\includegraphics[width=\textwidth]{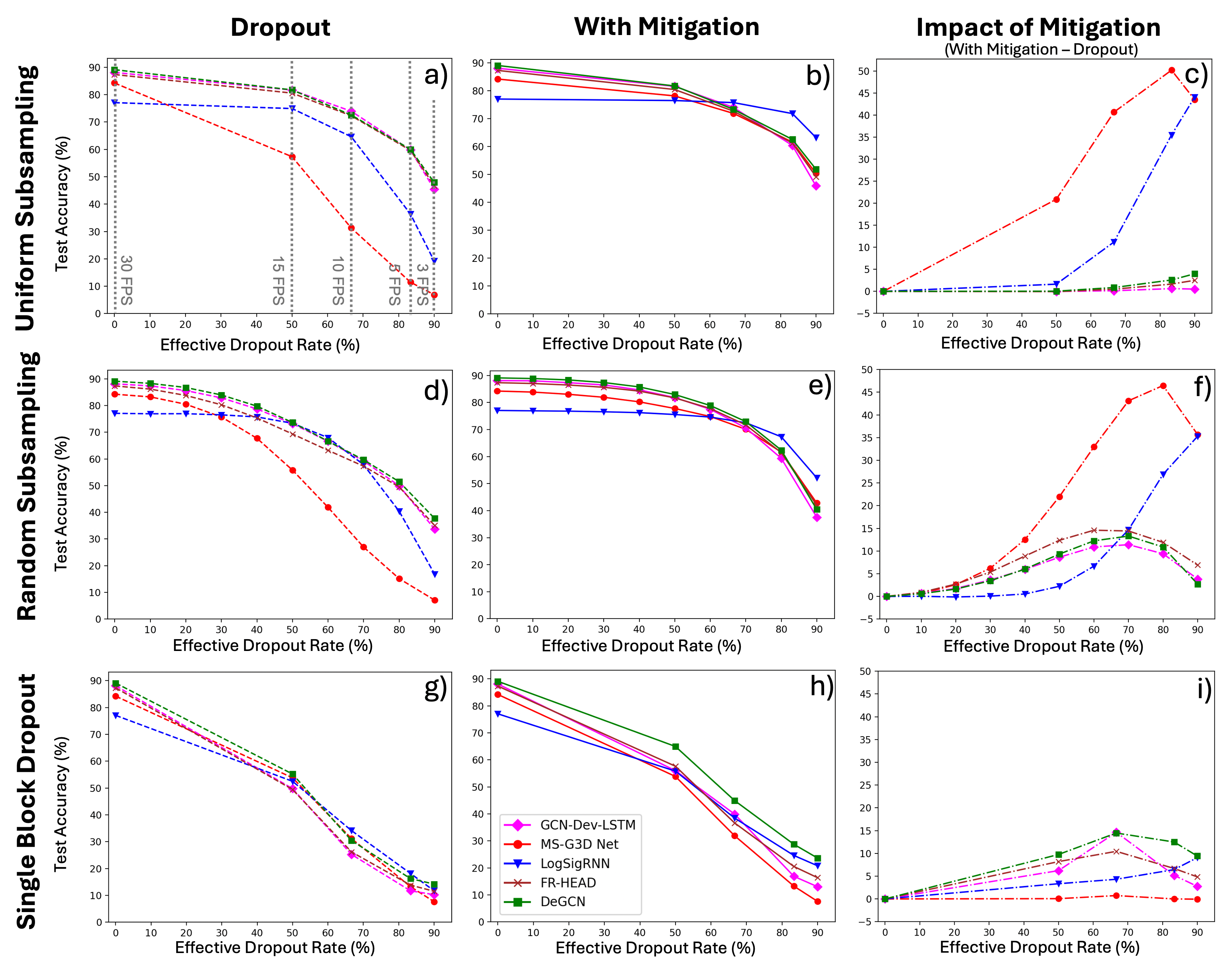}
\caption{Test accuracy versus effective dropout rate under the cross-setup train-test split for three different experiments, Uniform Subsampling ($a, b, c$), Random Subsampling ($d, e, f$) and  Single block Dropout ($g, h, i$) and five SHAR models (see key in panel $h$). The columns show experiments where removed frames are not replaced (left), where linear interpolation is used to mitigate degradation at inference time by replacing missing frames (central), and the difference (right). Panel $a$ also shows equivalent frame-rates in FPS for translation from effective dropout rate. All data are provided in Supplementary Tables 8-13.}
\label{fig:dropout_experiements_xset}
\end{figure}

Following the same experiments and layout from Figure 1 of the main paper, Figure \ref{fig:dropout_experiements_xset} shows the results of the 5 models under the 3 different experiments with and without mitigation. For the cross-subject train-test split in the main paper, we identified that for the single block dropout with mitigation the LogSigRNN model has a higher accuracy than the DeGCN model at $90\%$ effective dropout rate. But under the cross-setup train-test split at the same effective dropout rate, the results deviate slightly from the other five comparisons of DeGCN and LogSigRNN with mitigation; for the other five experiments LogSigRNN has higher accuracy than DeGCN by $1-12\%$, whereas for this one experiment DeGCN has $3\%$ higher accuracy than the LogSigRNN model (shown in Table 1, Figure 2 in the main paper and also Fig. \ref{fig:dropout_experiements_xset} h). We note however, that for this experiment the DeGCN and LogSigRNN accuracy/degradation curves have still converged with increasing magnitude of degradation, as for all other five experiments (see dotted green and blue lines in Figure 2, right panel in the main paper, or green and blue lines in Fig. \ref{fig:dropout_experiements_xset} h), even though they have not quite crossed at $90\%$ effective dropout rate. This leads us to suggest that that it is likely that the LogSigRNN will still outperform the DeGCN model effective dropout rates $>90\%$. Testing of such large dropout rates was not possible for this dataset, due to number of frames in some samples; at these very high levels of dropout there were too few frames remaining for some samples. We also expect that given the overall and consistently very low levels of performance for all models under the single block dropout, that the higher accuracy of DeGCN for this one experiment could be within error bounds; further tests over multiple random seeds would be required for clarification, which was beyond the computational resources available here. 

\end{document}